# Neural Probabilistic System for Text Recognition

Najoua Rahal, Maroua Tounsi, Adel M. Alimi

*REGIM-LAb. : REsearch Groups in Intelligent Machines, University of Sfax, National Engineering School of Sfax (ENIS), BP 1173, Sfax, 3038, Tunisia*

**Abstract**

Unconstrained text recognition is a stimulating field in the branch of pattern recognition. This field is still an open search due to the unlimited vocabulary, multi styles, mixed-font and their great morphological variability. Recent trends show a potential improvement of recognition by adoption a novel representation of extracted features. In the present paper, we propose a novel feature extraction model by learning a Bag of Features Framework for text recognition based on Sparse Auto-Encoder. The Hidden Markov Models are then used for sequences modeling. For features learned quality evaluation, our proposed system was tested on two printed text datasets PKHATT text line images and APTI word images benchmark. Our method achieves promising recognition on both datasets.

## 1. Introduction

The recognition of text, in both handwritten and printed forms, represents a fertile provenance of technical difficulties for OCR. Indeed, the printed is commonly governed by well established, calligraphy rules and the characters are well aligned. However, there is not always a system capable of reading printed text in an unconstrained environments such as unlimited vocabulary, multi size, mixed-font and their great morphological variability. This diversity complicates the choice of features to extract and recognition algorithm. The performance (speed and accuracy) of text recognition heavily relies on a robust features with rigorous recognizer that effectively fit variety while offering a great discriminative strength.

Usually, the input image is greatly correlated and thus can not be used for recognition as in its inherent form. The feature extraction step is crucial to eliminate the correlation from the original image. Given its importance, choosing appropriate features is difficult and requires a significant effort. Three main types of features are handcrafted in the area of the text recognition: 1) Transformation based features, 2) Structural features and 3) Statistical features. Transformation based features mutate the image from spatial representation to the required area. The structural features extract strokes, skeleton representation and loops from image. The statistical features focus on the analytic model like projection profiles, wavelets transforms coefficients, energy of Gabor filters and gradient histograms. The principal weakness in further of the *handcrafted* or heuristic features is that they are not robust and are computationally dense owing to great dimensions. In the recent decade, the new tendency in machine learning is the development of techniques that learn automatically script-independent features for the recognition of text recognition. In the last decade, the learned Bag of Features (BoF) becomes a greatly competitive representation. The main breakthrough that explain the strength of the BoF is the discriminative aspect of the low-level features which makes it a robust approach against the spatial variations.

The BoF framework was adopted in text recognition [**1**, **2**, **3**]. The methods mentioned previously illustrate that the BoF framework performs well in computer vision and pattern recognition branches. However, it suffers from a major problem which inhibits its efficiency.

In fact, the low-level visual dictionary, built using K-means, repudiates the spatial layout of the local descriptors. Thus, it brings a lack of the spatial information and therefore a shedding a discriminative power of separating the classes.

Applying to text recognition, neglecting the spatial patches information involves a confusion between character classes. Therefore, shapes of the same character must be discriminant.

We need a system to deal with the problems mentioned above by enhancing the power of the BoF image representation and improving the discrimination effectiveness. We propose, in the present paper, a new machine learning architecture of SAE for dictionary learning to benefit from the robustness of SAE in features representation.

Habitually, the most excellent word recognition systems are untrustworthy with a large number of words due to the unconstrained environment of the script. Furthermore, segmentation is a fundamental phase for any text recognition system. However, explicitly segmenting the text into recognition units such as characters or graphemes is too hard and complicated to implement. To avoid pre-segmentation problems, HMM, that have been efficiently used for text recognition, are opted. They influence, in a significative way, the performance and afterwards the recognition rate. Markov modeling has proved to be the classifier the most efficient and broadly used for text recognition. It is characterized by the avoidance of explicit segmentation of the words into grapheme, or character level. This relieves the errors of segmentation task. Moreover, HMM offer a stochastic modeling that struggles with variability of observation sequence lengths and nonlinear deformations, which perform them appropriate for the unconstrained environment of different script.

Our contributions are in the following manners:

1) A new SAE-based method is adopted to learn the visual dictionary than the traditional low-level visual dictionary built in a merely bottom-up way.

2) Our system leads to highly competitive performance, either surpassing the researches proposed in the literature on the PKHATT and APTI datasets.

3) Empirical studies indicate that using bigger visual dictionary yields superior recognition performance, which is prohibitively heavy and computationally expensive. The computation becomes unaffordable when a big dictionary is needful to fulfill the nonlinear manifold greatly. Contrariwise, the learning encoder-decoder networks, used in our work, is achieved by less wide visual dictionary, which is computationally much cheaper.

The rest of our paper is constructed as follows: In Section II, we discuss the methodology of our research work included SAE based BoF for feature extraction and HMM for recognition. The system is evaluated in Section III. The conclusion is drawn in Section IV.

## 2. Literature Review

While the great number of proposed approaches for printed Arabic text recognition, ago yet requires the improvement of accuracies in Arabic text recognition systems. This part provides a literature review covering HMM works for text recognition.

Bazzi *et al.* [4] investigated an unlimited-vocabulary and omnifont system for Arabic and English text line recognition. For feature extraction, the intensity, horizontal and vertical derivative of intensity, local slope and cor- relation across a window of two cells square are computed. The method is based on bigram and trigram language model for unlimited-vocabulary character recognition. As results, 3.3 % character error rate on mixed-font of 4 fonts (Geeza, Baghdad, Kufi, and Nadim) is achieved from the DARPA Arabic OCR Corpus, and 1.10% on Document English Image Database from the University of Washington.

Khorsheed [5] introduced a method for printed Arabic text recognition in line level. The main contribution is related to the use of discrete HMM. The pixel density features are extracted from cells falling into overlapped vertical windows. The proposed method was tested on a corpus containing 15,000 text line images written in 6 fonts in 600 format A4 pages. The obtained recognition rate is 95%. None peculiar handling for mixed-font recognition was presented.

Slimane *et al.* [6] presented an open vocabulary and multi-font printed Arabic text recognition from APTI database containing words in low resolution. The system consists of four main steps: font feature extraction, font recognition, word feature extraction and word recognition. For feature extraction, 102 features are extracted from each sliding window. The obtained feature vectors are then used to train the Gaussian Mixture Models based on the Expectation Maximization (EM) algorithm. At recognition step, an ergodic HMM model is used to permit the transitions between all the character models. The average recognition rates are 93.7% and 98.4%. The APTI database which is generated synthetically, was used for the evaluation of the developed system. But, the behaviour of the system was not studied in a more challenging task, i. e. the recognition on a real word images.

Prasad *et al.* [7] described an Arabic printed glyph recognition system based on HMM and Language Modeling. The glyphs are a conversion of the basis shape character transcriptions founded on the character form. The present work uses Position Dependent Tied Mixture (PDTM) HMM to model the different glyph representation models. The tree based state tying is used the position dependent training. For 176 characters, 339K Gaussians are trained. Although the previous work is considered as an efficient method, it presents few drawbacks. While the recognition of text in word level is acceptable, it is more suitable to recognize entire text lines rather than character or word levels. The recognition of entire text lines is beneficial since it allows to encompass every complication of segmentation and analysis of spaces between established words.

Natarajan *et al.* [8] presented a system recognition for Arabic, English, and Chinese scripts. Their contribution lies in the adoption of pixel percentile features which are robust to noise. Features are extracted from accumulated overlapped window cells. Also, angle and correlation from window cells are computed. The values at equal 20 divided pixel percentile (from 0 to 100) are attached to form a feature vector. Also, horizontal and vertical derivatives of intensity are also appended. They prove the efficiency of their features to recognize text from different scripts.

Al-Muhtaseb *et al.* [9] described an unlimited-vocabulary printed Arabic text recognition technique for 8 fonts; Akhbar, Andalus, Arial, Naskh, Simplified Arabic, Tahoma, Traditional Arabic, and Thuluth. The novelty in this system is the adoption of sliding window in a hierarchical structure. 16 features are extracted from vertical and horizontal overlapping and non-overlapping windows. The obtained average accuracies of recognition vary between 98.08% and 99.89% for the eight fonts.

Ait-Mohand *et al.* [10] proposed an interesting method of mixed-font text recognition using HMMs. The principal contribution of the method was associated to HMM model lengthiness adaptation techniques which were incorporated with maximum likelihood linear regression (MLLR) and maximum a posteriori (MAP) techniques. The proposed techniques were efficient in recognition of text in mixed-font. This method suffers from two limitations: each font requires small amounts of data to be evaluated, also, during the evaluation the line images would be obtained from a single font.

Ahmad *et al.* [11] presented mono-font and mixed-font Arabic printed text recognition based HMM. In this work, the features are extracted from adaptive sliding windows. The text is recognized at line level from 8 fonts: Akhbaar, Andalus, DecoType Thuluth, Naskh, Tahoma, Traditional Arabic, Simplified Arabic and Times New Roman. The proposed approach includes two phases. In the first phase, the font of the input text line is identified. In the second phase, the HMM is trained on the associated font for recognition. The obtained results show the efficiency of the proposed method for mono- font. The achieved recognition rates are between 98.96% and 92.45% CRR. By cons, this method suffers from limited results with mixed-font, about 87.83%.

The previously detailed methods presented a successful application of HMM to provide segmentation free in mixed-font printed text recognition. As HMM are used in different contexts, the difference is highlighted in the enhancement of features representation resulting in a high recognition rate. However, the results are still yet to be improved. A trustworthy and powerful system remains a very challenging task and a main objective of pattern recognition researches. In the present work, we focus on the exploitation of SAE based BoF for feature extraction to enhance and perform the recognition task.

## 3. SAE based BoF for Text Recognition

The use of SAE for codebook learning has shown its robustness and its ability to capture the high level content of an image. This demon- strates its power against image clutters and occlusions. The codebook generated by SAE has proven its effectiveness for classification and recognition in unconstrained environment. Applied to text recognition, the experiments show the strength of SAE that resides on its robustness to the mono font and mixed font recognition. It fits well with the elusive notion of similarity that includes: the diversity between instances of the same character and the similitude between instances of different character categories. Our chosen of SAE for codebook generation has a huge impact on the recognition performance. The reason for using SAE is also its representation learning method that can unsupervisedly encode data into a new representation while exploiting its spatial relations. Also, it has the aptness to learn complicated nonlinear re- lationships and the robustness according to the data provided, and produces better results.

Hereby in the present part, we detail the proposed system for text recognition. After normalization, it starts with the Dense SIFT low level features extraction in the input image. At the learning stage, a visual dictionary is created using an unsupervised SAE. Afterwards, features are computed with respect to the learned visual dictionary. Thereafter, given that HMM model sequence of feature vectors, a sequence of BoF representations is generated by sliding a vertical window along the image in writing direction. Histograms of SAE based BoF are presented as the input extracted features of the HMM. In HMM, the system realizes the training and recognition stages. Training process starts with the feature vector sequences and the corresponding transcription files to produce the character models. In the recognition process, the SAE BoF feature vectors are presented to a network of HMM character models. A word can be described by a concatenation of character models. **Figure 1** below describes the outline of our system and presents the tasks that are listed above.

### 3.1. Low Level Feature Extraction

We start with a preprocessing step, in which the images are normalized to a fixed height whereas keeping the aspect ratio. As shown in **Figure 1(B)**, Dense SIFT [**12**] are computed over a 4 x 4 spatial grid into 8 bins. While different detectors have been suggested in computer vision, such as Harris, Harris Laplace, Hessian-Laplace, SURF, however, DSIFT by patch based descriptors have the most splendid performances for text recognition [**13**, **14**].

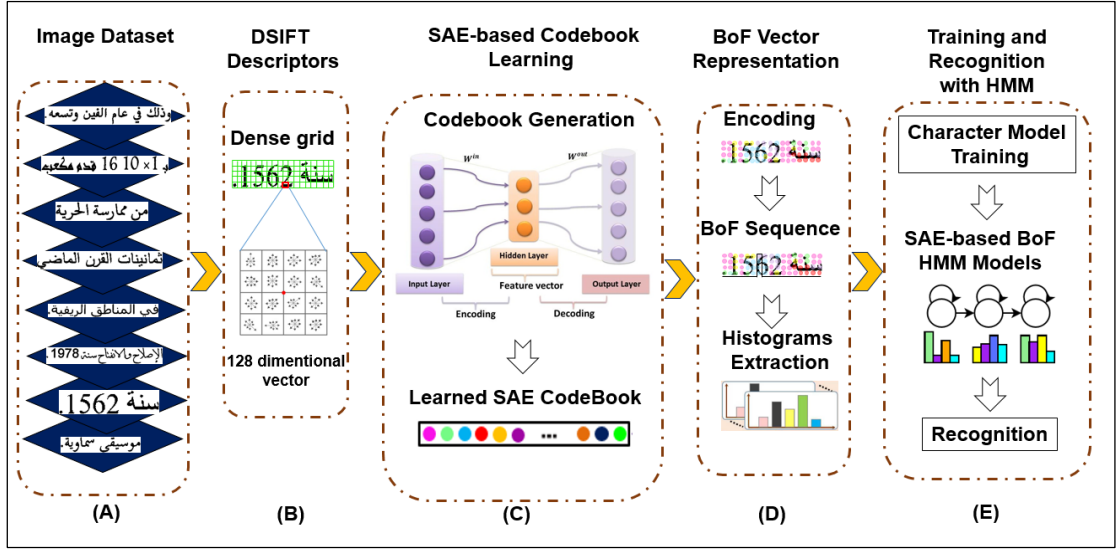

**Figure 1:** SAE based BoF for Text Recognition System

## 3.2. SAE Visual Dictionary Learning

As shown in **Figure 1(C)**, in this section, we build a hidden layer using the SAE in order to learn the visual dictionary and fine-tune the discriminative feature representations for each class character. SAE is a vigorous unsupervised and non-linear learning neural network used for mapping function training.

It insures the reconstruction error minimization between the input and the reconstruction data. Therefore, SAE learns, in a low dimensional space, an efficient representation, which consists of encoding decoding. Encoding of an input data $X$ with size $N$ is achieved by a non-linear activation function:

$$Z = f(W^1 X + b_1) \qquad (1)$$

Where $W^1 \in \Re^{L \times K}$ are a set of weights and $b_1 \in \Re^K$ is the encoding bias vector. The decoding of $Z$ is achieved using the following matrix:

$$\hat{X} = f'(W^2 Z + b_2) \qquad (2)$$

where $W^2 \in \Re^{K \times L}$ and $b_2$ represent the decoding matrix and bias respectively.

The parameters $W^1$, $W^2$, $b_1$ and $b_2$ are learned by the following reconstruction error minimizing:

$$J(W^1, W^2, b_1, b_2) = 1/2N \sum_{i=1}^{N} \|\hat{x}_i - x_i\|_2^2 + \lambda \|W^1 + W^2\|_2^2 \qquad (3)$$

The *sparse* AE includes a sparsity constraint minimizing the Kullback Leibler (KL) divergence [15] on hidden layer activation. That is to say, we append a regularization idiom to the reconstruction error in (3):

$$J(W^1, W^2, b_1, b_2) + \beta KL(p\|\hat{p}_j) \qquad (4)$$

where $\beta$ verifies the weight of the sparsity penalty. $p$ is the target average activation placed to the nearest centroid for data points lower a confidence threshold. $\hat{p}_j$ is the average activation of the hidden layer. The KL divergence is defined by:

$$\beta KL(p\|\hat{p}_j) = p \log p/\hat{p}_j + (1-p) \log 1 - p/1 - \hat{p}_j \qquad (5)$$

### 3.3. Vector Image Representation

As shown in **Figure 1(D)**, the visual dictionary computed in the previous step is used to quantize the descriptors in a word image. Afterwards, since the HMM are a stochastic model which produces a sequence of observations, they need a sequence of BoF representations. For this purpose, a word image is divided into frames and the sequence of BoF vectors are obtained by sliding a vertical window shifted from right to left direction. Finally, we get a set of histograms as a vector image representations.

### 3.4. Training and Recognition with HMM

The HMM modeling is implemented using the toolkit HTK (Hidden Markov Models Toolkit) [16]. It operates in two tasks: training and recognition. In both tasks, the same extracted features are used. Each image is transformed to a sequence of SAE based BoF feature histograms. For data preparation, the sequence of feature histograms are converted into a compatible format with HTK using the *HCopy* tool. *HCompV* tool is then used to estimate the overall mean and variance vectors. The training data is used for HMM character models initialisation. We employed Linear HMM topology, in which only self and next state probabilities are taken with fixed number of states per model, as illustrated in **Figure 2**.

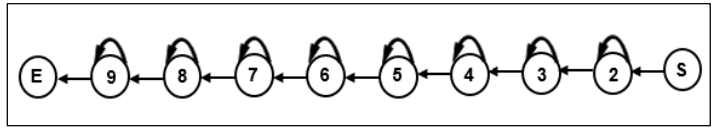

**Figure 2:** 9 states right to left HMM for modeling characters. The starting state S and ending state E are non-emitting states.

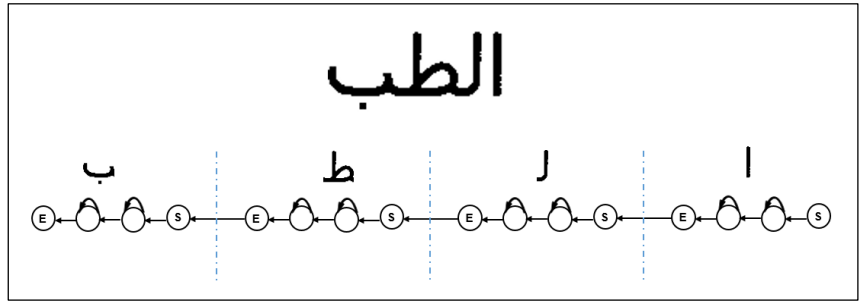

**Figure 3:** Example of character sub-models of a word

An HMM is formed by the connected character sub-models of a word, as shown in **Figure 3**.

In the training task, the Baum-Welch iterative estimation process is implemented with the tool *HREst* to refine the parameters Gaussian mixtures, means and variances.

At the recognition task, extracted features from testing images are tied in HMM character models. An ergodic HMM allows the transition between character models and therefore the recognition in an open vocabulary. The recognition is achieved by providing the great sequence states of character based on Viterbi algorithm launched by *HVite* tool. Thus, the obtained recognition results are converted to Unicode Characters. They are evaluated in term of Character and Word recognition rates obtained by *HResult* tool. The HTK tool *HResult* reported the performance of our system in terms of Line Recognition Rate (LRR), Recognition Rate (CRR) rates and Character Word Error Rate (WER), which take into consideration the errors due to substitution, insertion and deletion.

For complete text image recognition, an ergodic HMM is used to concatenate the models. In ergodic HMM topology, each character model may be achieved from any other character model in a finite number of transitions. Thanks to this topology, it allows the recognition in unlimited vocabulary. An example of ergodic HMM with 4 character models is illustrated in **Figure 4**.

## 4. Experiments and Results

In order to assess the performance and generality of our system, two datasets are used to show results across multiple data types, as detailed in subsection 4.1. In order to be able to evaluate our experiments leaded in mixed-font text line level, we use P-KHATT dataset to compare our results to [17]. To demonstrate the efficiency of our system in several databases, we present a comparative study with other systems tested using the Arabic Printed Text Image (APTI) database [18]. Then, in subsection 4.2, we describe our experiments. Finally, we present, in subsection 4.3, a detailed evaluation of our system.

### 4.1. Datastes

#### 4.1.1. P-KHATT Dataset

P-KHATT [17] contains text lines from eight fonts: *Akhbaar, Andalus, Naskh, Simplified Arabic, Tahoma, DecoType Thuluth, Times New Roman and Traditional Arabic*. The text is scanned at resolution 300 dots/inch. It includes the staple 28 Arabic letters with their different shapes and combinations, Space, 10 digits and punctuations ('.', ',', ':', ';', '!', '(', ')', '?', '-', '/', '%', etc). It contains 6472 text line images for training, 1414 text line images for development and 1424 text line images for testing. For another information about the total number of text lines, words and characters in P-KHATT, we assign [17]. A few samples of this dataset are shown in **Figure 5**.

#### 4.1.2. APTI Dataset

APTI [18] consists of a word images generated in 10 fonts (Arabic Transparent, Tahoma, Andalus, AdvertisingBold, Simplified Arabic, Traditional Arabic, Diwani Letter, M Unicode Sara, Naskh, and DecoType Thuluth), 10 font sizes (6, 8, 10, 12, 14, 16, 18 and 24 points), and 4 font styles (plain, bold, italic, and combination of italic bold). It contains 113,284 words synthetically generated in low resolution with 72dpi. APTI is elected for the evaluation of OCR systems. Some images are shown in **Figure 6**.

### 4.2. Configurations of Text Recognition System

#### 4.2.1. Parameters for Preprocessing and Features Extraction

Our system involves four processing steps : i) preprocessing, ii) feature extraction with SAE-BoF, iii) training and iiii) recognition. **Table 1** summarizes the key configuration of our system for preprocessing and feature extration steps. In the preprocessing step, the normalization height of 55 pixels, while observing the width aspect ratio, allows to reduce the side effect of the variety in font size. As shown in **Table 2**, 55 pixels height is likely to give best recognition. For feature extraction with DSIFT, the image spatial area is divided into a grid of overlapping fixed-sized where the descriptors are extracted from each patch. For feature extraction with DSIFT, the descriptors are extracted from each patch.

The SAE architecture used in this work is formed by one Auto Encoder (AE). Feeding the latent representation (*c*) of the AE, where the input of the AE is the original data features.

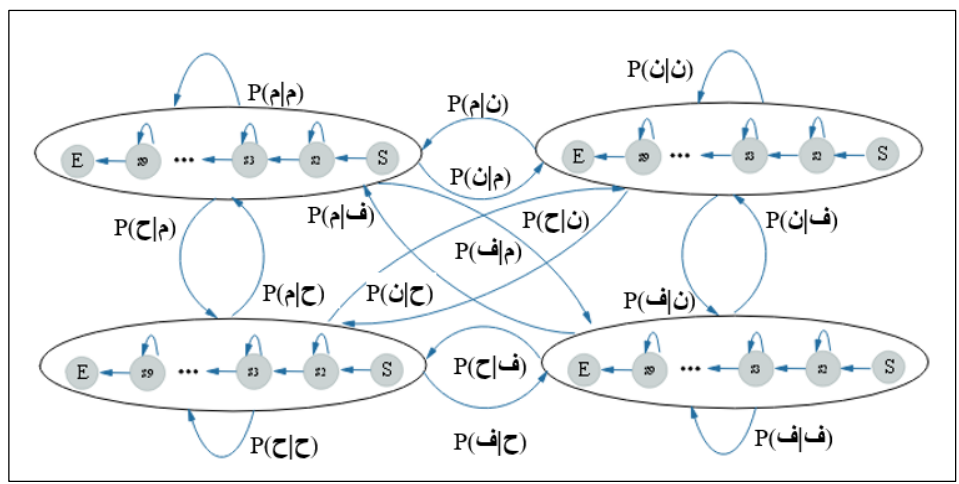

**Figure 4:** Ergodic HMM topology composed by 4 character models

في اتخاذ القرارات للصالح العام. : A

في اتخاذ القرارات للصالح العام. : B

في اتخاذ القرارات للصالح العام. : C

في اتخاذ القرارات للصالح العام. : D

في اتخاذ القرارات للصالح العام. : E

في اتخاذ القرارات للصالح العام. : F

في اتخاذ القرارات للصالح العام. : G

في اتخاذ القرارات للصالح العام. : H

**Figure 5:** Samples of PKHATT Dataset

البحث : A          البحث : F
البحث : B          البحث : G
البحث : C          البحث : H
البحث : D          البحث : I
الحث : E           البحث : J

**Figure 6:** Samples of APTI Dataset

### 4.2.2. Training and Recognition with HTK

For training and recognition, we exploited the HTK. In the first step, the BoF representation are taken from each text image and the transcription file is prepared. In the training step, firstly, the HMM model of each character is initialised using the HTK tool *HCompV*. Secondly, the Baum-Welch re-estimation is done in several iterations using the HTK tool *HERest*.

This phase involves an extra step applied to increase the Gaussian mixtures number. Then, the recognition is done by looking for the maximum probability of each character model using the Viterbi algorithm executed with the tool *HVite*. Finally, an ergodic HMM is used to concatenate character models. The HTK tool *HResult* reported the performance of our system in terms of Line Recognition Rate (LRR) and Character Recognition Rate (CRR) rates, which take into consideration the errors due to substitution, insertion and deletion.

Guided by [**19**], the number of character models has a strong effect on the performance of recognition. They proposed 10 systems with different set of character models : $Set_{35}$, $Set_{36}$, $Set_{38}$, $Set_{42}$, $Set_{61}$, $Set_{62}$, $Set_{64}$, $Set_{120}$, $Set_{124}$, $Set_{68}$. In the present work, we apply $Set_{62}$ which leads to the best results. All parameters, previously detailed, are set based on the recognition results for the Tahoma font of P-KHATT database.

## 4.3. Experiments

Various experiments are applied to evaluate the statistical features extracted by the proposed SAE-BoF framework. Also, we concentrate on the evaluation of parameters for HMM characters modeling. The experiments are conducted on mono-font and mixed-font text recognition. Like in [**17**], for Tahoma font recognition, 2000 text lines were used for the training and 1424 text lines for the testing.

For mixed-font recognizer, the set of training and the testing text lines from the eight fonts were employed.

### 4.3.1. Impact of Varying Patch and Stride Size

In the implementation of proposed text recognition system, there are a few significant parameters combined with text recognition performance, notably, the variation of the number of patch and stride size for the BoF. We draw samples of three patches : 3 x 3 with 1 stride, 5 x 5 with 2 strides and 8 x 8 with 4 strides. The best values for the patch size and stride were generated based on the recognition results for the Tahoma font of P-KHATT database.

**Table 3** reports the results of our strategy for the P-KHATT database. We observe that our system provides the best accuracy recognition rate of 99.95% at character level and 99.40% at text line level for a patch size P=5 and stride D=2. Herein, it is interesting to notice that more we increase the patch size, more the recognition rate decreases.

Considering that the best performance was obtained with histogram features parameterized with P=5 and D=2, we keep these configurations for the following described experiments.

### 4.3.2. Impact of Varying Window and Shift Size and States Number

**In Table 4**, we studied different sliding window widths (W) and Shifts (S) with several number of states to find the best values of character models. The optimal accuracies are obtained with parameters W=4, S=3 and Number of States=10.

### 4.3.3. Effect of Hidden Layer Size of SAE

The feature learning layer represents the hidden layer of SAE. The neurons number in the hidden layer represents the size of the SAE codebook. Different hidden neurons number (50, 250 and 500) are adopted to test its impact on the SAE learning. **Table 5** shows that the SAE codebook size affects the recognition accuracy. We observe that the proposed text recognition system provides its best performance with dictionary size = 500. We obtain an average accuracy of 99.95% for CRR and 99.40% for LRR.

**Table 1:** Configuration of our Text Recognition System

| Parameter | Value |
|---|---|
| Preprocessing | Height Normalization =55 |
| Descriptor | DSIFT |
| AE | Hidden Size : 500<br>L2WeightRegularization : 0.1<br>SparsityRegularization : 1<br>SparsityProportion : 0.95<br>Loss Function : msesparse |

**Table 2:** Impact of Varying Height Size

| Height Size | CRR(%) | LRR(%) |
|---|---|---|
| 45 | 99.91 | 98.25 |
| 55 | 99.95 | 99.40 |
| 60 | 99.92 | 98.00 |

### 4.3.4. Impact of the SAE Codebook Generation

SAE represents more flexibility then the hard clustering algorithm like K-means. The obtained results, as shown in Table 6, demonstrate a better performance of SAE to those derived via K-means.

With regards to the mono-font and mixed-font, the codebook generated with SAE reaches the best performance than the codebook generated with K-means. Furthemore, for Tahoma mono-font, a best average text recognition accuracy of 99.95% for CRR is achieved with SAE. It raises an improvement up to 3.25% comparing to k-means codebook.

### 4.3.5. Impact of different number of Gaussian Mixtures

The basic profit of the Gaussian Mixtures is their power to model complicated shapes of functions of probability density. They are modeled more precisely with increasing the Gaussians number. **Figure 7** demonstrates the growth of the CRR and LRR rates as a function of the Gaussians number. Wehighlight that the CRR and LRR rates of Tahoma font text recognition are respectively increased from 97.90% and 70.35% with 1 Gaussian to 99.95% and 99.40% with 64 Gaussians. As indicated on the curve evolution, going more than 64 Gaussians does not ensure the enhancement of the obtained results.

**Table 3:** Impact of Patch and Stride Size

| Patches/Strides | CRR(%) | LRR(%) |
|---|---|---|
| 3/1 | 99.88 | 98.20 |
| **5/2** | **99.95** | **99.40** |
| 8/4 | 99.69 | 93.41 |

**Table 4:** Variation of Text Recognition Rates by varying the window and shift Size and States Number

| W/S | Number of States | CRR(%) | LRR(%) |
|---|---|---|---|
| 2/0 | 7 | 74.78 | 41.06 |
| 3/2 | 10 | 99.87 | 97.80 |
| 4/1 | 6 | 99.76 | 95.10 |
| **4/3** | **10** | **99.95** | **99.40** |
| 5/2 | 6 | 99.79 | 95.60 |
| 6/3 | 6 | 99.77 | 95.20 |

**Table 5:** Different Visual Dictionary sizes
**PKHATT**

| Hidden Layer Size | CRR (%) |
|---|---|
| 250 | 99.90 |
| **500** | **99.95** |
| 700 | 99.93 |

**Table 6:** Compariosn of Text Recognition Rates via K-means and SAE

| Codebook Generation | Tahoma | | Mixed-Font | |
|---|---|---|---|---|
| | CRR(%) | LRR(%) | CRR(%) | LRR(%) |
| K-means | 96.70 | 95.18 | 94.70 | 85.33 |
| SAE | 99.95 | 99.40 | 98.92 | 90.00 |

## 4.4. Comparison to State of the Art

In **Tables 7** and **8**, we compare our results with state of the arts methods. The comparison proves the effectiveness and the robustness of our system. In fact, the system proposed by [**17**] is validated on the P-KHATT database. The comparison is assessed using mono-font and mixed-font. The extracted features with SAE show their contribution in solving the problem of morphological differences between the characteristics of characters belonging to different fonts. They allow a global and robust parametrization whatever the case of mono-font context or mixed-font context. The recorded performances are very promising in recognition of Arabic in unconstrained environments. The results are pre-sented in Character Error Rate (CER) idiom.

In **Table 9**, we further compare our system to existing HMM based systems that are reported in the literature using APTI database. The Arabic Transparent is the font dependable in the comparison such that it is closed to the reference protocols for the text recognition competitions using the APTI database. [**20**, **21**].

**Table 10** presents an evaluation of multi-font mono-size systems. Tests are performed using "Arabic Transparent" Plain and sizes 24, 18, 12, and 10. The best results are obtained by our system. The obtained rates declare the stability of our system to the variability of sizes. A fully literal comparison is yet not feasible for multiple reasons.

One of the further important reason is that set6, which is not publicly available, is used only to evaluate the systems in the competitions. For other systems that used the APTI database and that are available in the literature, each system built its particular training, development, and evaluation set. A few systems applied word lexicons and n-gram LM, whereas other systems did not use them. P-KHATT and APTI database share a fixed parameters of the overall system.

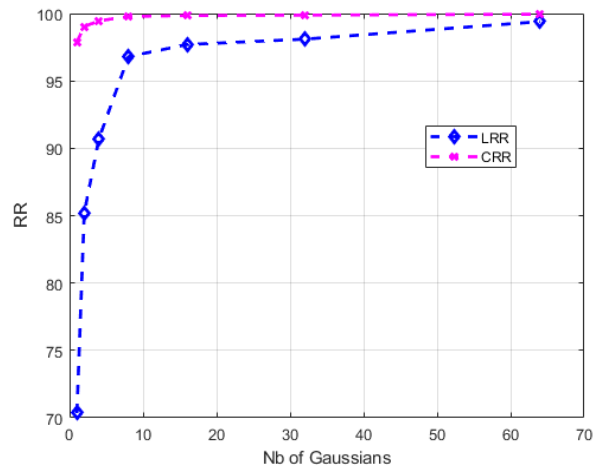

**Figure 7:** Growth of Recognition Rate (RR) by improving the number of Gaussians

**Table 7:** Comparison of mono-font recognition
**PKHATT**

| Font | CER (%) | |
| --- | --- | --- |
| | Ahmad *et al.* [17] | SAE-BoF-HMM |
| Times New Roman | 1.20 | **0.1** |
| Andalus | 1.35 | **0.04** |
| DecoType Thuluth | 7.55 | **0.29** |
| Tahoma | 1.04 | **0.05** |
| Traditional Arabic | 4.35 | **0.23** |
| Naskh | 3.06 | **0.24** |
| Akbaar | 2.87 | **0.35** |
| Simplified Arabic | 1.67 | **0.04** |

**Table 8:** Comparison of mixed-font recognition
**PKHATT**

| System | CER (%) | |
| --- | --- | --- |
| | Mixed-font using samples from all fonts | Mixed-font using font identification |
| **Ahmad *et al.* [17]** | 12.19 | 3.44 |
| **SAE-BoF-HMM** | **1.08** | - |

**Table 9:** Comparison with HMM Text Recognition
**APTI**

| Features Description | Training | Testing | CER (%) | WER (%) |
|---|---|---|---|---|
| DIVA-REGIM [20] | Sets{1-5} | Set6 | 0.30 | 1.10 |
| Black and White connected components and their ratio, compactness, gravity centre, horizontal and vertical projection | | | | |
| UPV-BHMM [22] | 10.000 | 3.000 | 0.30 | 1.70 |
| Black and White connected components | | | | |
| THOCR2 [20] | Sets{1-5} | Set6 | 0.81 | 4.97 |
| Structural and Statistical features and their derivatives | | | | |
| Awaida and Khorsheed [23] | 80.000 | 14.418 | 3.35 | - |
| Run-Length Encoding (RLE) | | | | |
| Ahmad et al. [17] | Sets{1;2} | Set5 | 0.57 | 2.12 |
| Statistical features and their derivatives | | | | |
| **SAE-BoF-HMM** | **Sets{1;2}** | **Set5** | **0.05** | **0.17** |
| SAE based BoF | | | | |

**Table 10:** Evaluation of the Multi-Size Mono-Font System
**APTI**

| System/Size | | 8 | 10 | 12 | 18 | 24 | Mean RR |
|---|---|---|---|---|---|---|---|
| IPSAR [20] | WRR | 73.3 | 75.0 | 83.1 | 77.1 | 77.5 | 65.3 |
| | CRR | 94.2 | 95.1 | 96.9 | 95.7 | 96.8 | 98.7 |
| UPV-PRHLT-REC1 [20] | WRR | 97.4 | 96.7 | 92.5 | 84.6 | 84.4 | 91.7 |
| | CRR | 99.6 | 99.4 | 98.7 | 96.9 | 96.0 | 98.3 |
| DIVA-REGIM [20] | WRR | 95.9 | 95.7 | 93.9 | 97.9 | 98.9 | 94.8 |
| | CRR | 99.2 | 99.3 | 98.8 | 99.7 | 99.7 | 99.1 |
| **SAE-BoF-HMM** | **WRR** | **99.1** | **99.2** | **99.2** | **99.5** | **99.5** | **99.3** |
| | **CRR** | **99.9** | **99.9** | **99.9** | **99.9** | **99.9** | **99.9** |

## 5. Conclusion and Future Work

This paper is about the recognition of unlimited-vocabulary Arabic text images. A good analysis of literature review in Arabic text recognition lead us to develop a novel system to handle some of the current challenges. More specifically, the challenges of unlimited-vocabulary and mixed-font are addressed. We use BoF framework based on SAE for codebook generation. It enhances the quality of the extracted features and produces a robust statistical features for holistic text recognition system. Herein, the main advantage of BoF based on SAE is their ability to cope with the font identification where each input text is associated with the closet known font. Also, they demonstrate their ability to handle the specifities of fonts, showing complex shapes with ligatures and overlaps with the same parameters used to handle an easy font without ligature or overlap. Several critical parameters of the BoF are exeprimentally evaluated including the patch, the stride and the codebook size. In other words, we investigated the selection of HMM for training and recognition. An interesting feature of HMM is in the implicit segmentation that the system is able to perform automatically. This feature is especially interesting for Arabic script where a priori segmentation of the characters is hard due to the cursive nature of the scripts (being printed or handwritten). So, HMM are able to perform the segmentation and recognition simultaneously. The obtained results in this paper are encouraging future works. The recognition of text with degraded and highly complex documents, such as administrative documents, could be investigated.

## Acknowledgment

The research leading to these results has received funding from the Ministry of Higher Education and Scientific Research of Tunisia under the grant agreement number LR11ES48.